\documentclass[10pt,twocolumn,twoside]{IEEEtran}
\IEEEoverridecommandlockouts

\usepackage{cite}
\usepackage{amsmath,amssymb,amsfonts}
\usepackage{algorithmic}
\usepackage{graphicx}
\usepackage{textcomp}
\usepackage{balance}
\usepackage{xcolor}
\usepackage[utf8]{inputenc}

\usepackage{eurosym}

\usepackage{array} 
\usepackage{multirow}
\usepackage{booktabs}  
\usepackage{siunitx}   
\usepackage{caption}   
\usepackage{enumitem}

\newcolumntype{C}[1]{>{\centering\arraybackslash}m{#1}}
\newcolumntype{L}[1]{>{\raggedright\arraybackslash}m{#1}}

\newcommand{\mediqr}[2]{\num{#1} $\pm$ \num{#2}}
\newcommand{\ci}[2]{[\num{#1}, \num{#2}]}

\setlist[itemize]{nosep,leftmargin=*}
\setlist[enumerate]{nosep,leftmargin=*}

\newcolumntype{L}[1]{>{\raggedright\arraybackslash}p{#1}}

\usepackage{tikz}

\def\BibTeX{{\rm B\kern-.05em{\sc i\kern-.025em b}\kern-.08em
    T\kern-.1667em\lower.7ex\hbox{E}\kern-.125emX}}
\usepackage{lipsum}
\usepackage{url}

\title{\LARGE \bf 
A Comprehensive Framework for Long-Term Resiliency Investment Planning under Extreme Weather Uncertainty for Electric Utilities 
}

\author{Emma Benjaminson$^\dagger$
\vspace{-0.5cm}
 \thanks{ 
$^\dagger$Emma Benjaminson is with the Boston Consulting Group's AI Science Institute, Pittsburgh, PA, USA. Email: {\tt benjaminson.emma@bcg.com}}
}

\begin{document}
\begingroup
\allowdisplaybreaks

\maketitle

\begin{abstract}
Electric utilities must make massive capital investments in the coming years to respond to explosive growth in demand, aging assets and rising threats from extreme weather. Utilities today already have rigorous frameworks for capital planning, and there are opportunities to extend this capability to solve multi-objective optimization problems in the face of uncertainty. This work presents a four-part framework that 1) incorporates extreme weather as a source of uncertainty, 2) leverages a digital twin of the grid, 3) uses Monte Carlo simulation to capture variability and 4) applies a multi-objective optimization method for finding the optimal investment portfolio. We use this framework to investigate whether grid-aware optimization methods outperform model-free approaches. We find that, in fact, given the computational complexity of model-based metaheuristic optimization methods, the simpler net present value ranking method was able to find more optimal portfolios with only limited knowledge of the grid.
\end{abstract}


\section{Introduction}\label{intro}

A paradigm shift is currently underway in the design and operation of electric utility grids worldwide. Utility operators must commit to massive investments in growth, modernization and resilience efforts in the next few years in response to a rapidly changing natural and economic environment - in the US alone, utilities are collectively expected to invest over \$1 trillion dollars between 2025 and 2029 \cite{utilitydive2025}. Broadly, there are 3 key factors driving this shift: 1) an exponential increase in demand driven by data centers and the electrification of industry and transportation; 2) aging infrastructure in need of replacement, and 3) shifting extreme weather patterns increasing in frequency and severity, necessitating grid hardening activities to improve grid resilience \cite{utilitydive2025, lbnlbrattle2025}. In the face of these mounting pressures, utilities must respond proactively while still providing electricity safely, reliably, and equitably.


With regards to planning long-term investments for improving grid resilience - an essential activity for utilities in this moment - utilities already have a robust, multi-step process, for example as detailed in NREL's reports on resilience planning \cite{nrelhurricanes, nrelwildfires, nrelwinterstorms}. These reports identified a central framework for resilience investment planning, beginning with identifying hazards, defining key metrics for evaluating investment plans, then quantifying risks posed by hazards as well as the potential benefits from improvements under consideration, before finally developing an optimal portfolio and schedule of investments. The NREL reports, as well as researchers like Baker \cite{baker2019} and Parker et al. \cite{parker2023}, also emphasize that resilience efforts must not cement inequitable systems for the future. 

Recent approaches for optimizing capital investments for resilience, ranging from prioritization frameworks like TOPSIS \cite{niu2023} to MILP formulations \cite{yang2021, xue2024} to evolutionary algorithms \cite{yankson2025, wu2025} tend to cover a subset of the steps described in the NREL process, but do not yet present a comprehensive end-to-end solution. We also noted that some methods in the literature use a model of the grid, while others are more model-free, primarily focused on the financial or operational benefits of different investments. This presents an interesting question: is there a measurable benefit to leveraging a detailed model of the grid while optimizing capital investments for resilience?

In addition to studying the value of a model-aware investment optimization approach, the NREL studies also identified a need for utilities to use advanced methods for multi-objective optimization under uncertainty \cite{nrelhurricanes, nrelwildfires, nrelwinterstorms}. Guided by this recommendation, we propose a novel framework for end-to-end capital investment optimization for resilience, and use it to investigate the potential benefits of grid-aware optimization as compared to model-free methods. The framework includes: 
\begin{enumerate}
    \item A probabilistic approach for modeling extreme weather
    \item A digital twin of the grid, including all generators, power lines and loads, as well as a simulator capable of solving the AC power flow problem at steady state
    \item A method for capturing uncertainty in grid performance 
    \item A multi-objective optimization approach capable of finding an optimal combination of resilience improvements 
\end{enumerate}

We demonstrate an implementation of this framework, and use it to test our hypothesis that a grid-aware approach for optimizing a portfolio of investments will find synergistic network effects in the grid that cannot be identified using a model-free approach. We test our hypothesis by applying both a model-free net present value ranking method and a grid-aware metaheuristic multi-objective optimization approach to define an optimal portfolio of investments. We use the digital twin of the grid to evaluate both optimized investment plans, and compare them across several key performance metrics to identify the benefits and drawbacks of each method. 
\vspace{-10pt}

\section{Literature Review}
Prior art has explored a range of optimization approaches to prioritize capital investments, starting with interpretable ranking approaches that score projects across criteria. Nematshahi et al. ranked projects according to probability-weighted wildfire damage costs \cite{nematshahi2024} and Niu et al. used the Technique for Order of Preference by Similarity to Ideal Solution (TOPSIS) method for ranking projects according to a weighted score of operational, social and economic criteria \cite{niu2023}. Another group of papers used mixed integer linear programming (MILP) approaches. Yang et al. used a MILP formulation to perform multi-objective optimization given constraints regarding project sequencing, complementarity and mutual exclusivity \cite{yang2021}. Xue et al. also derived a MILP formulation for a problem that was originally defined as a two-stage optimization problem, starting with siting and sizing for grid-level energy storage, and then solving the day-ahead clearing problem for power markets. They collapsed the two-stage problem to one layer by using the KKT conditions and the linearized DistFlow model \cite{xue2024, vonmeier2024}. Several approaches used an evolutionary algorithm, like Yankson et al. who used the Improved Grey Wolf Optimizer to find optimal groupings of resources to form microgrid structures \cite{yankson2025}. Wu et al. introduced a new evolutionary algorithm, super-efficient hyperplane projection transformation (which is based on NSGA-III), to define a Pareto frontier \cite{wu2025}.

\begin{figure*}[t]
    \centering
    \includegraphics[width=0.93\textwidth]{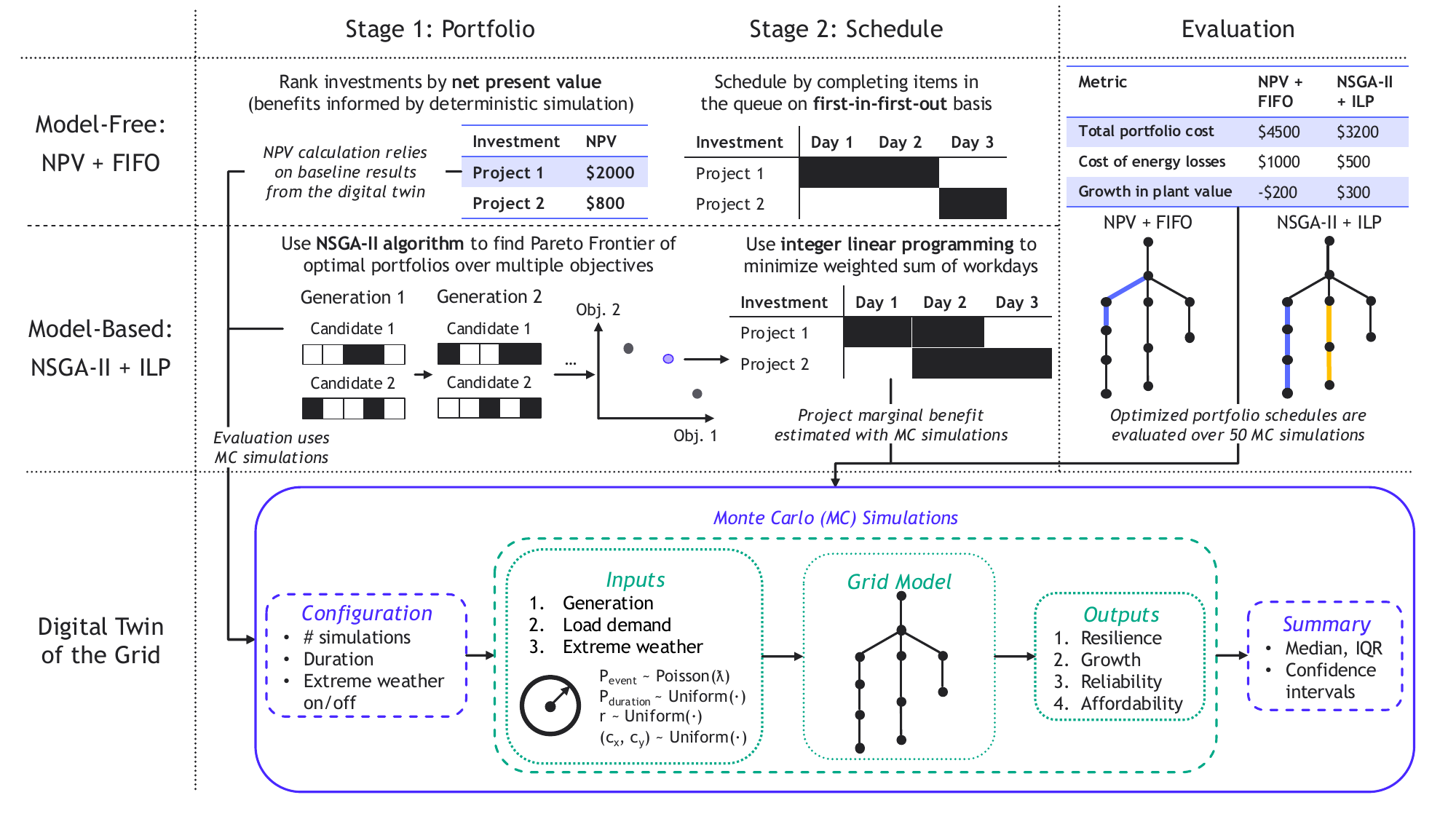}
    \caption{This work compares a model-free and a model-based approach to solving a two-stage capital prioritization problem for electric utilities. \textbf{Model-free approach:} The potential investments are ranked by NPV, which is calculated using a baseline simulation from the digital twin to estimate operational costs and benefits, and then scheduled on a FIFO basis. \textbf{Model-based approach:} The NSGA-II algorithm \cite{deb2002} finds the Pareto frontier of optimized portfolios over two objectives, using the digital twin as an evaluation function. An ILP method is used to optimize the schedule of implementing the investments. \textbf{Digital twin of the grid:} The digital twin computes power flows over the grid layout given known generation, load and extreme weather events. Multiple simulations can be run to obtain summary statistics and confidence intervals for KPIs across different categories including resilience, growth and affordability.} 
    \label{fig:main_flow}
    \vspace{-10pt}
\end{figure*}

Several papers also accounted for sources of uncertainty, such as weather and load growth, in their solution. Nematshahi et al. used a fire spread simulator to model the spread and intensity of a fire over time under varying conditions. The results were used to compute a weighted average of the damage cost given the probability of different conditions \cite{nematshahi2024}. Similarly, Donadel used Monte Carlo simulation to study the impact that demand variability (modeled as a probability distribution) would have on the value of different reconductoring plans \cite{donadel2025}. Xue et al. also used a probability distribution to capture the uncertainties in wind and solar generation as deterministic constraints based on a Gaussian mixture model-based representation \cite{xue2024}. 


Previous works have also leveraged grid models to varying extents, suggesting that grid-aware optimization can be advantageous. Works by Meinecke et al. \cite{simbench}, Yankson et al. \cite{yankson2025}, Donadel \cite{donadel2025} and Xue et al. \cite{xue2024} all use power flow modeling in their implementations. Leveraging power flow modeling for either representative scenarios (Yankson et al. performed N-1 impact analysis), or over time (Meinecke et al., Donadel and Xue et al. all had long-term planning horizons) was crucial to these approaches' solutions. Other methods surveyed that did not directly mention performing power flow analyses still relied on grid-related metrics. For example, Yang et al. \cite{yang2021} and Niu et al. \cite{niu2023} leveraged grid metrics related to voltage qualification rate, capacity-load ratio, and N-1 pass rate that could have been calculated using power flow analysis. And even papers that were less clearly reliant on a detailed model of the grid still leveraged information like the grid topology \cite{nematshahi2024} and operations \cite{wu2025}. 


Finally, it is important that we ensure resilience efforts do not preserve systems of energy inequity \cite{baker2019}. One step towards transformative change is to develop metrics that clearly measure energy inequities today and our progress towards addressing them. As Parker et al. explain, many of these metrics will need to include socioeconomic data, for example to compute energy burden, which is a ratio of the annual household energy expenditure to the annual household income \cite{parker2023}. There will also need to be a shift towards de-averaging metrics, so that instead of computing the SAIDI, SAIFI and CAIDI for all of the utility's customers, we compute these metrics over a smaller region, while taking into account customers' sociodemographics \cite{parker2023}. 

\section{Methods}
We introduce a novel four-part framework for finding the optimal resilience investment plan in the face of uncertainty. We will use this framework to investigate our research question: does a grid-aware optimization approach outperform a model-free method when searching for an optimal resilience investment plan? We will explore this question by comparing a model-based, multi-objective optimization approach, which can query the model of the grid during the optimization process, to a model-free approach that ranks investments according to net present value. Both methods' optimized investment plans are evaluated using the digital twin, and then compared across multiple metrics.


Our test cases are synthetic distribution grids, and the set of possible resilience investments includes burying overhead power lines to reduce outages due to high winds, and upgrading power lines to increase their maximum current rating and reduce resistive losses. There is limited open-source data available for time series datasets for US-style distribution systems \cite{datasetsreview} so we used the SimBench time series datasets for subsets of the German grid from low voltage (distribution) to extremely high voltage (transmission) grids \cite{simbench}. The time series data are synthetic representations of 2016 profiles for different generation and load sources \cite{simbench}.  
\vspace{-10pt}
\subsection{Probabilistic Model of Extreme Weather}\label{methods_extreme_weather_patterns}
We define a probabilistic model of an extreme weather event, in particular a thunderstorm. We represent the probability that the storm will occur as a Poisson process whose number of events, $\lambda$, in a given time interval (1 hour) is dependent on the season and time of day. Specifically, $P_{event} \sim \text{Pois}(\lambda)$, where $\lambda = \text{base probability per hour} \cdot \text{seasonal multiplier} \cdot \text{hourly multiplier} \cdot \text{timestep interval}$. Thus, if $P_{event} > 0$, then we initialize an extreme weather event. The base, seasonal and hourly multipliers are set to indicate that thunderstorms occur more frequently in the afternoons between summer and winter. All multipliers are in Tables \ref{test_case_1_values} and \ref{test_case_2_values} in the Appendix.

We also represent the event's impact center, impact radius, and duration as probability distributions. The impact center, $(c_x, c_y)$, is sampled from a pair of uniform distributions, bounded by the geographic range of the grid model:
\begin{align*}
    c_x & \sim U[x_{min}, x_{max}) \\
    c_y & \sim U[y_{min}, y_{max})
\end{align*} 

Similarly, the impact radius and the duration of the event are also sampled from uniform distributions. 

If the sampling from the Poisson distribution determines that an extreme weather event occurs, the simulation identifies power lines within the event's impact area and samples from a uniform random distribution to determine if they will be impacted. Once a line goes down, it is not returned to active status in the simulation until the event is over and the repair time has elapsed. We estimate the repair time as a function of the line length and type (i.e. overhead or underground). 
\vspace{-10pt}
\subsection{Digital Twin of the Grid}
For detailed modeling of extreme weather events, the digital twin of the grid must represent the operation of the grid over an investment planning time horizon with fine granularity. The digital twin can be used by a grid-aware optimization method to compute the value of the objective function given a proposed investment plan. Specifically, the digital twin computes the power flow within the grid at 15 minute intervals, leveraging Pandapower \cite{pandapower} to perform the actual power flow calculation, with extreme weather events included as described above.

The power flow results provide information about losses, line congestion and the actual power production at the slack bus, which can then be used to compute the key performance metrics that inform the objective function. We also track outages due to extreme weather events, and report corresponding metrics including unserved electricity and SAIDI-MED, SAIFI-MED and CAIDI-MED (see Table \ref{performance_metrics}).
\vspace{-7pt}


\subsection{Modeling Uncertainty}\label{mc_sims}
In this work, uncertainty is introduced by the probability of an extreme weather event striking some part of the grid. We take a Monte Carlo approach to modeling the uncertainty, which involves running many scenarios where the storm occurrence and impact region will vary. Each scenario has the same time series dataset for load demand and power generation, but stochastic weather events. Once the set of scenarios has been simulated using the digital twin of the grid, we can compute summary statistics and confidence intervals (using the t-distribution for situations with $<$30 scenarios, and the z-distribution for $\geq$30 scenarios) for each of our KPIs to quantify the most likely performance characteristics of the grid. 

\subsection{Approach for Optimizing Resilience Investment Portfolio}\label{optimize_for_resilience}
We solve a two-layer optimization problem in this work. The upper-level problem is finding the optimal portfolio of investments that fit within the constraints. The lower-level problem is determining the optimal schedule for completing the investments. The model-agnostic method uses net present value (NPV) ranking for the upper-level problem and first-in-first-out (FIFO) ordering for the lower-level scheduling problem; for a grid-aware approach, we use non-dominated sorting genetic algorithm II (NSGA-II) \cite{deb2002} for the portfolio optimization and an integer linear programming (ILP) approach for solving the scheduling problem. In order to evaluate both methods, we will conduct the same set of Monte Carlo simulations over each method's optimized investment plan, and compare them using the metrics in Table \ref{performance_metrics}. 

To demonstrate how equity could be incorporated in our metrics, we report the optimized investment plans' KPIs both for the entire grid as well as for each subnet (i.e. neighborhood). De-averaging the KPIs would be the first step in reviewing the optimized plans' impacts on individual communities, and in a real world setting (as opposed to using synthetic data as we do in this work), overlaying other information such as median household income could help build a clearer picture of how the utility's planned investments would improve metrics such as energy burden by neighborhood \cite{parker2023}. This information would then inform discussions between the utility and the community to ensure alignment on the planned investments. 


For the upper-level problem of optimizing the portfolio of investments, the objectives for the model-based, multi-objective approach are to minimize unserved energy (caused by outages due to extreme weather events) and to minimize energy losses due to resistive heating. The decision variables include undergrounding overhead power lines and upgrading underground power lines to a conductor with a higher maximum current rating. The constraint is the total budget. The objective of the lower-level problem of scheduling the investments is to minimize the weighted completion time (where weights prioritize the projects with the largest marginal benefit). The decision variables are the technician assignments that must be optimized to complete all projects. The constraints are a limited number of available technicians.

\subsubsection{Method 1: NPV Ranking with FIFO Scheduling}
We define the NPV of an investment as: 
\begin{equation*}
    NPV = \sum_{t=0}^T \frac{(B_t - C_t)}{(1 + r)^t}
\end{equation*}
Where $B_t$ is the benefit during the time period $t$, $C_t$ is the cost during $t$, $r$ is the discount rate and $T$ is the asset lifetime. 

The annual benefits, $B_t$, of our investments include the reduction in cost of the total unserved energy (for investments that bury power lines to reduce outages) and the reduction in costs due to resistive heating losses (for all investments). We determine the unserved energy and resistive losses by comparing a simulation of the grid where the upgrade is applied to a simulation where no changes are made. We estimate the unserved energy costs by assuming that the overhead line is down for one day in the summer (the same day and duration is used for every power line). This is a simplifying assumption that the investment in undergrounding will prevent one power line outage per year, and that there would be exactly one outage per year in the base case. 

Similarly, for computing losses due to resistive heating, we will compare the total losses for the two line ratings over the year (assuming no extreme weather events). We estimate the operations and maintenance costs, $C_t$, per year of the line as 10\% of the initial capital investment cost. 

Once the NPV is calculated for every investment, we rank them and select as many as will fit within the budget constraint; projects that have both negative NPV and a ratio of $\frac{|NPV|}{B_t} > 1.1$ are excluded by default. The investments are then scheduled on a FIFO basis, where we approximate the time required to complete each project as a function of the length of the line, and we complete as many projects in parallel as possible as long as they are adjacent in the queue, and while there are technicians available.  

\subsubsection{Method 2: NSGA-II \& ILP Scheduling} 
Here we use the NSGA-II algorithm to solve the portfolio optimization problem and an ILP formulation to solve the subsequent scheduling problem. The NSGA-II algorithm is implemented using pymoo \cite{pymoo} and the ILP problem is solved using pyomo \cite{hart2011pyomo, bynum2021pyomo}. The NSGA-II is configured to find a set of optimal solutions along the Pareto frontier over the two objectives. Each potential solution is represented as a binary-valued vector, where each value corresponds to a particular investment. Standard point crossovers and bitflip mutations from pymoo are used to evolve the population \cite{pymoo}. 

For each candidate solution, the evaluation function runs multiple Monte Carlo simulations over the year 2016 using the digital twin of the grid, and returns the average values of the optimization objectives computed over the set of simulations. We penalize constraint violations by quantifying how far over the allowed budget and/or timeline a candidate solution goes. We implemented parallelization and caching of individual candidate solutions for greater computational efficiency. Even with these enhancements, the runtime for our NSGA-II algorithm on our test cases took multiple days, and so we note that future work should include exploring further opportunities for acceleration, such as using a surrogate model for faster evaluation of candidate solutions. Additional algorithm parameters are listed in Tables \ref{test_case_1_values} and \ref{test_case_2_values}. 

\begin{figure}[t]
    \centering
    \includegraphics[width=0.5\textwidth]{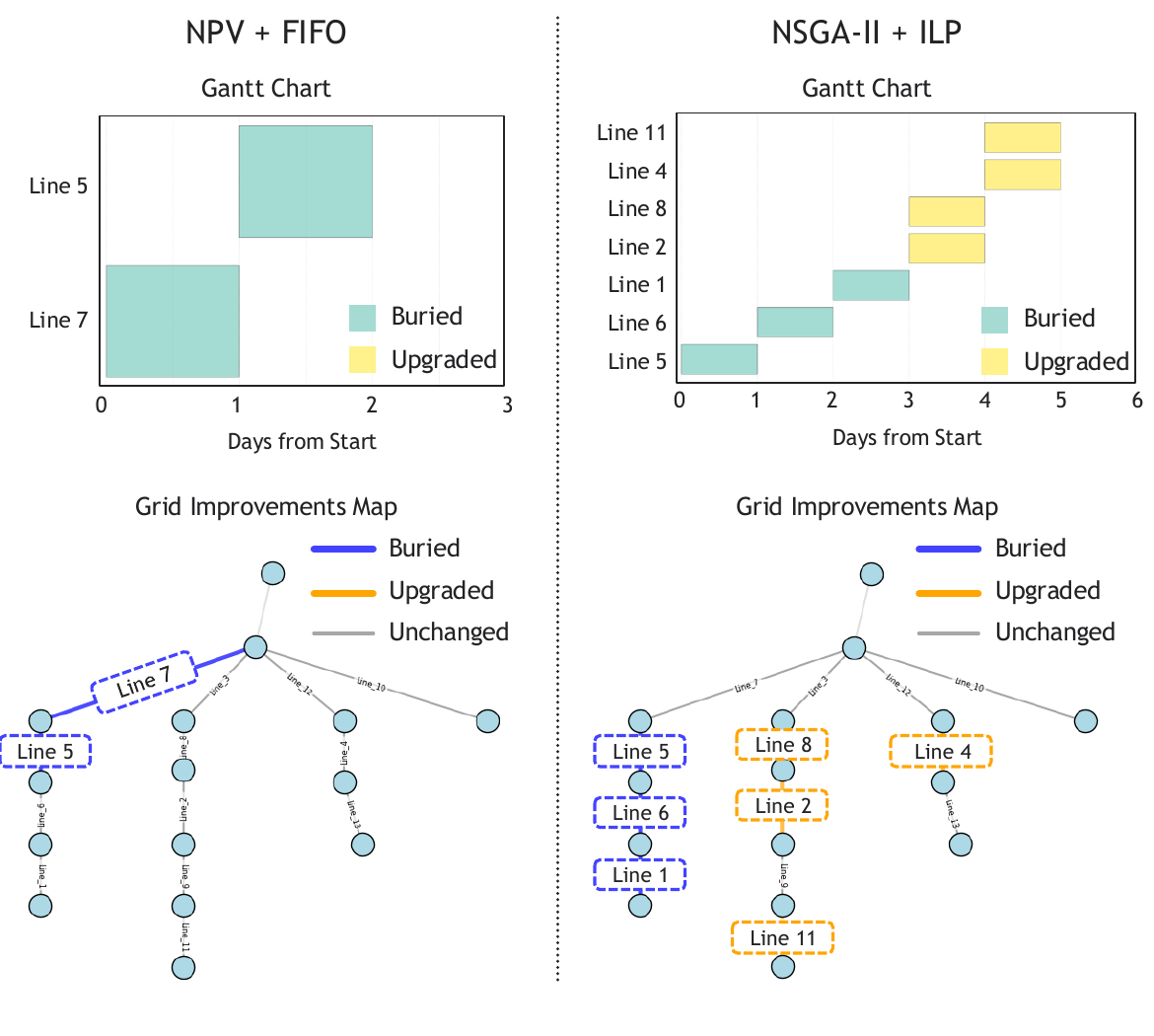}
    \caption{\textbf{Test case 1 results:} Comparison of optimized portfolio and schedule for model-free  vs model-based approaches. \textbf{NPV + FIFO:} Lines 7 and 5 are buried, addressing the two most critical lines on the left-most feeder branch. \textbf{NSGA-II + ILP:} Gantt chart ordering prioritizes projects with the largest marginal benefit. Portfolio includes both burying some overhead lines (not line 7) and upgrading others.}
    \vspace{-15pt} 
    \label{fig:test_case_1_results}
\end{figure}

\begin{table*}[t]
\centering
\caption{Results for Test Case 1 (Varying with Extreme Weather)}
\label{test_case_1_results}
\begin{tabular}{
  L{4.0cm}
  C{2.5cm} C{3.25cm}
  C{2.5cm} C{3.25cm}
}
\toprule
\textbf{Metric} &
\multicolumn{2}{c}{\textbf{NPV + FIFO}} &
\multicolumn{2}{c}{\textbf{NSGA-II + ILP}} \\
\cmidrule(lr){2-3}\cmidrule(lr){4-5}
& \textbf{Median $\pm$ IQR} & \textbf{95\% Confidence Interval} &
  \textbf{Median $\pm$ IQR} & \textbf{95\% Confidence Interval} \\
\midrule
Cost of total unserved energy [\$] & \mediqr{1310}{1320} & \ci{1130}{1700} & \mediqr{1450}{1410} & \ci{1290}{1890} \\
SAIDI-MED [min]                   & \mediqr{496}{512}   & \ci{456}{646}   & \mediqr{569}{452}   & \ci{518}{717} \\
SAIFI-MED                         & \mediqr{1.08}{1.06} & \ci{1.01}{1.39} & \mediqr{1.38}{0.865}& \ci{1.24}{1.67} \\
CAIDI-MED [min]                   & \mediqr{459}{172}   & \ci{408}{470}   & \mediqr{421}{115}   & \ci{377}{439} \\
\bottomrule
\end{tabular}
\vspace{-10pt}
\end{table*}

\begin{table}[h]
\centering
\caption{Results for Test Case 1 (Fixed Across All Scenarios)}
\label{test_case_1_results_fixed}

\begin{tabular}{
  L{4.0cm}
  S[table-format=5.0]
  S[table-format=5.0]
}
\toprule
\textbf{Metric} & \textbf{NPV + FIFO} & \textbf{NSGA-II + ILP} \\
\midrule
Total investment portfolio cost [\$]      & \bfseries 3720  & 17900 \\
Total number of investments               & 2               & 7 \\
Total days to complete work               & 2               & 5 \\
Cost of total resistive losses [\$]       & 994             & \bfseries 494 \\
Growth in plant value [\$]                & -531            & \bfseries 15300 \\
Average capacity headroom [\%]            & 97.2            & \bfseries 97.4 \\
Duration above capacity limit [\%]        & 0               & 0 \\
Duration of voltage deviation [\%]        & 0               & 0 \\
\bottomrule
\end{tabular}
\vspace{-15pt}
\end{table}

Once the Pareto frontier is identified using the NSGA-II algorithm, we can select an optimized portfolio and solve the second stage problem. In this experiment we choose the portfolio with the best score over all objectives. The decision variables are: 
\begin{align*}
    x[i, j, t] &\in \{0, 1\} \\ 
    C[i] &\in \mathbb{Z}, \quad 0 \leq C[i] \leq N_{projects} \\
    y[i, t] &\in \{0, 1\} \\
    &\forall i \in N_{projects}, j \in N_{technicians}, t \in N_{days} \notag
\end{align*}
Where $x$ is the set of project assignments for the technicians, $C$ is the list of project completion dates and $y$ is a set of indicators tracking which projects are active on which date. The number of projects is set by the chosen portfolio and the number of technicians is fixed by the test case. The number of days allocated to the total schedule is equal to the sum of days required to complete all projects sequentially. The objective is to minimize the weighted sum of the completion dates for all projects: 
\begin{equation*}
    \min \sum_i^{N_{projects}} w[i] \cdot C[i]
\end{equation*}
Where $w[i]$ is the marginal benefit for the $i$-th project. We can estimate the marginal benefit for each project by comparing the portfolio's performance without that project compared to the complete portfolio's performance. The marginal benefit is the financial value of the reductions in unserved energy and resistive losses enabled by the project in question. We estimate the value of unserved energy and resistive losses by running Monte Carlo simulations with extreme weather included. Finally, our ILP formulation has four key constraints:
\begin{enumerate}
    \item Project duration for each project must be met by the sum over the technician assignment variable $x$.
    \begin{equation*}
        \sum \big( x[i, j, t] \text{ } \forall j, t \big) = \text{duration}[i]
    \end{equation*}
    \item Technicians can only work on one project per day.
    \begin{equation*}
        \sum \big( x[i, j, t] \text{ } \forall i \big) \leq 1
    \end{equation*}
    \item The completion day must be last day of work on project.
    \begin{equation*}
        C[i] \geq t \cdot x[i, j, t]
    \end{equation*}
    \item Projects will need a fixed number of technicians to work on them simultaneously.
    \begin{equation*}
        \sum \big( x[i, j, t] \text{ } \forall j \big) = \text{technicians required}[i] \cdot y[i, t]
    \end{equation*}
\end{enumerate}

\section{Results and Discussion}
\begin{table*}[t]
\centering
\caption{Results for Test Case 2 (Varying with Extreme Weather)}
\label{test_case_2_results}

\begin{tabular}{
  L{4.2cm}
  C{2.5cm} C{3.25cm}
  C{2.5cm} C{3.25cm}
}
\toprule
\textbf{Metric} &
\multicolumn{2}{c}{\textbf{NPV + FIFO}} &
\multicolumn{2}{c}{\textbf{NSGA-II + ILP}} \\
\cmidrule(lr){2-3}\cmidrule(lr){4-5}
& \textbf{Median $\pm$ IQR} & \textbf{95\% Confidence Interval} &
  \textbf{Median $\pm$ IQR} & \textbf{95\% Confidence Interval} \\
\midrule
Cost of total unserved energy [\$M] & \mediqr{2.10}{2.01} & \ci{2.22}{4.06} & \mediqr{2.12}{1.98} & \ci{2.21}{4.05} \\
SAIDI-MED [hr]                   & \mediqr{24.9}{25.9}   & \ci{24.9}{34.9}   & \mediqr{24.6}{23.7}   & \ci{24.6}{34.6} \\
SAIFI-MED                         & \mediqr{0.934}{0.613} & \ci{0.879}{1.12} & \mediqr{0.939}{0.613} & \ci{0.875}{1.11} \\
CAIDI-MED [hr]                   & \mediqr{27.4}{12.9}   & \ci{26.3}{30.6} & \mediqr{26.9}{12.4}   & \ci{26.0}{30.4} \\
\bottomrule
\end{tabular}
\vspace{-10pt}
\end{table*}

\begin{table}[t]
\centering
\caption{Results for Test Case 2 (Fixed Across All Scenarios)}
\label{test_case_2_results_fixed}

\begin{tabular}{
  L{4.0cm}
  S[table-format=5.0]
  S[table-format=5.0]
}
\toprule
\textbf{Metric} & \textbf{NPV + FIFO} & \textbf{NSGA-II + ILP} \\
\midrule
Total investment portfolio cost [\$]      & \bfseries 450\,000  & 467\,500 \\
Total number of investments               & 2                   & 4 \\
Total days to complete work               & 25                  & 29 \\
Cost of total resistive losses [\$]       & \bfseries 196\,000  & \bfseries 196\,000 \\
Growth in plant value [\$]                & -187\,000           & \bfseries -142\,000 \\
Average capacity headroom [\%]            & \bfseries 95.9      & \bfseries 95.9 \\
Duration above capacity limit [\%]        & 0               & 0 \\
Duration of voltage deviation [\%]        & 0               & 0 \\
\bottomrule
\end{tabular}
\vspace{-15pt}
\end{table}

\subsection{Test Case 1: Small Rural Network}
This test case is drawn from \texttt{1-LV-rural1--0-no\_sw} in SimBench \cite{simbench}. It contains 13 lines, 14 buses, one transformer and a connection to the external grid. The original base configuration defined all lines to be underground, and so we modified lines 1, 5, 6 and 7 to be overhead lines of type \texttt{94-AL1/15-ST1A 0.4}, which is compatible with the nominal voltage of the system and the closest to the underground lines in cross-sectional area and maximum current rating. Additional parameters are listed in Table \ref{test_case_1_values}.

The NPV + FIFO method found there were only two investments (lines 5 and 7) that had positive values. After FIFO scheduling was implemented, all projects could be completed in 2 days, well within the time constraint, and the total project cost ($\$3\,720$) was much less than the allocated budget (see Table \ref{test_case_1_results_fixed}). As shown in Figure \ref{fig:test_case_1_results}, lines 5 and 7 were closest to the external grid and after they were upgraded, the simulation results showed that the downstream overhead lines on the same feeder branch (lines 1 and 6) experienced the majority of outages, which we would expect given that they were the two remaining vulnerable overhead lines. However, after observing that the SAIDI-MED and CAIDI-MED were also relatively high (see Table \ref{test_case_1_results}), it suggested that our modeled extreme weather probabilities might be higher than typically seen in reality, and so we decreased them for the second test case. We also noted that the growth in plant value was negative because the total depreciation of the grid dominated the investments made to the two overhead lines. And we observed substantial capacity headroom and no excursions above capacity and voltage limits, likely because the load profiles were selected to present normal conditions and not high congestion conditions. 
\vspace{-1pt}
\indent In contrast, the NSGA-II + ILP method selected a total of seven investments, leading to a much larger portfolio cost than the NPV approach. The investments led to a large positive growth in plant value and a 50\% reduction in resistive losses. However, the median values for the cost of unserved energy and the SAIDI-MED and SAIFI-MED were higher than for the NPV + FIFO method. This is due to the fact that the NSGA-II algorithm did not select the critical line 7 as a portfolio investment, and so the entire branch was left vulnerable to more frequent outages. The NSGA-II algorithm may not have identified the critical importance of line 7 because it was limited to 3 generations and a population size of 5 candidate solutions, so the algorithm may never have seen the effect of upgrading line 7, or perhaps not frequently enough for it to be selected in subsequent generations. We were limited in the generations and population size by the compute time necessary to run each Monte Carlo simulation over the entire year at high granularity. We did note that, while the NSGA-II algorithm did not identify the central importance of line 7, the ILP solution was able to prioritize the undergrounding of lines 5, 6 and 1 in the order in which they are positioned along the feeder line, so that the upstream lines were completed first. This indicates that the scheduling approach was able to account for the impact of potential line outages once the optimal portfolio was identified. 
\vspace{-10pt}

\subsection{Test Case 2: Midsize Commercial Network}
Our second test case is drawn from \texttt{1-MV-comm--2-no\_sw} in SimBench \cite{simbench}. It contains 113 lines (both overhead and underground) organized into 6 subnets, 107 buses, 2 transformers and 1 connection to the external grid. The search space presented in this test case is substantially larger than that presented in Test Case 1 (considering each line as a potential investment). Given the substantial compute time required to test each candidate solution in the NSGA-II algorithm, we chose to apply a heuristic to prioritize a subset of lines to consider for investment. Based on the findings from Test Case 1, especially the importance of lines directly connected to power sources, we selected all of the lines connected to the external grid or to other large ($\geq 1$MW) generators (with the exception of buried lines 103 and 104 that were already using the largest available conductor material). 

We also sought to expand the search conducted by the NSGA-II algorithm by increasing the population size to 8 candidates and the number of generations to 5, and the number of Monte Carlo simulations per candidate from 3 to 6. However, in order to facilitate this expansion, we had to reduce the simulation time span from a full year to 90 days. While this does not capture full seasonal variation, it does provide some insight into the impact of outages induced by extreme weather to give an approximate measure of the value of different investments. In future work, finding further computational speedups and training a surrogate model would be essential to speeding up the simulation time of the digital twin to enable more complete exploration of the search space. Test case parameters are provided in Table \ref{test_case_2_values}. 

The NPV + FIFO approach invested in burying two of the three available overhead lines, prioritizing the two lines that were connected to radial paths, lines 8 and 39, instead of loops (line 19), see Figure \ref{fig:test_case_2_results} in the Appendix. The NSGA-II + ILP approach selected lines 8 and 19, not line 39, for undergrounding, as well as two other lines for conductor upgrades. Again, even though we had expanded the population size and number of generations for this iteration of the NSGA-II algorithm, we still found that it failed to identify a critical investment (line 39) that would provide substantial resilience benefits. However, as shown in Table \ref{test_case_2_results}, the cost of total unserved energy and other parameters across both methods were highly comparable because we still only applied investments to a relatively small set of power lines, and outages on other overhead lines continued to dominate this metric. We also noted in Table \ref{test_case_2_results_fixed} that although the NSGA-II + ILP method upgraded the conductor on 2 power lines that were not in the NPV + FIFO portfolio, they made an inconsequential impact on the cost of total resistive losses. 

Finally, we evaluated our two optimized portfolios on the cost of total unserved energy at the subnet level, as shown in Figure \ref{fig:test_case_2_subnets}. We can see that our portfolios had differing impacts on subnets 3 and 5, providing a more granular view of the impact of our portfolio optimization. 

\begin{figure}[h]
    \centering
    \includegraphics[width=0.5\textwidth]{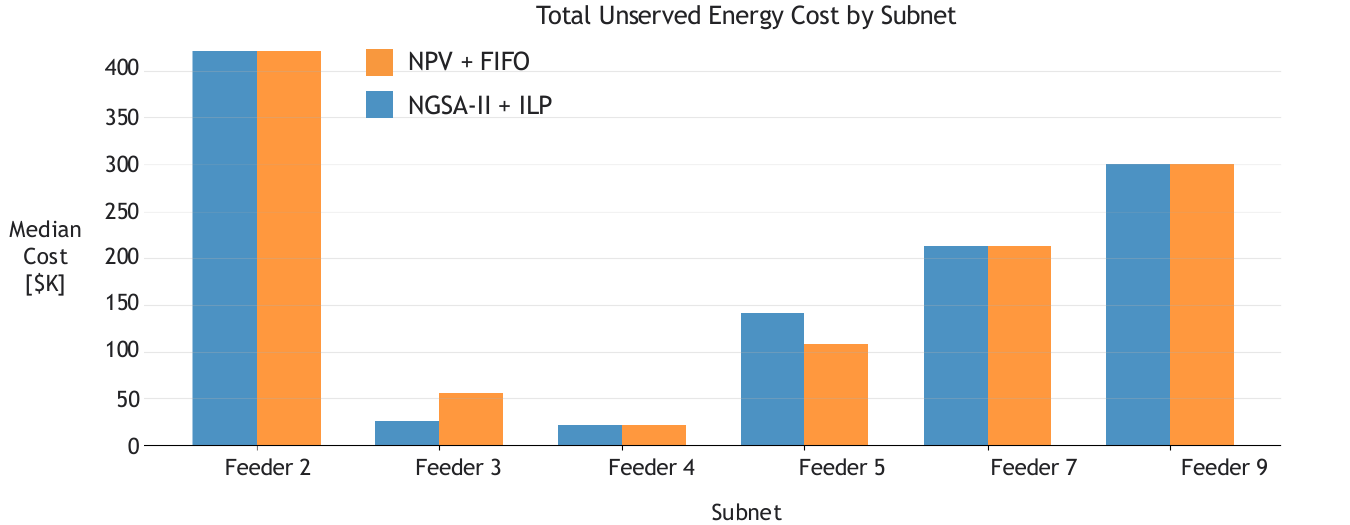}
    \caption{\textbf{Test case 2 results: }The NPV + FIFO portfolio reduced the median unserved energy in Feeder 5 subnet as compared to the NSGA-II + ILP portfolio, and vice versa for Feeder 3.} 
    \label{fig:test_case_2_subnets}
    \vspace{-10pt}
\end{figure}

\section{Conclusion}
This work investigated the value of using grid-aware optimization methods for capital prioritization for electric utilities. We found that the NPV approach consistently invested in the most critical lines, while the NSGA-II algorithm was not always able to identify them. However, the grid-aware ILP scheduling method was able to more intelligently prioritize work on investments as compared to the FIFO method. The NSGA-II algorithm is computationally expensive and this limitation prevented it from fully exploring the search space. Thus, while there might be an advantage to having more complete access to a model of the grid, we found that the NPV method with limited access to the digital twin was sufficient to provide an optimal solution. In future work, we will facilitate faster evaluation of the digital twin, possibly using surrogate models to generate approximate solutions.

\section{AI Usage Disclosure}
The authors used AI to search for relevant literature, but not for paper summarization once articles were selected. AI was used to write code for data processing and visualization, for developing the digital twin of the grid, for implementing the optimization approaches and for plotting results. Experimental design and paper writing were completed entirely without AI. 

\newpage
\appendix 
\begin{table}[ht]
\centering
\caption{Performance metrics used to evaluate optimization approaches}
\label{performance_metrics}

\begin{tabular}{L{1.5cm} L{2.0cm} L{4.0cm}}
\toprule
\textbf{Category} & \textbf{Metric} & \textbf{Definition} \\
\midrule

\multirow{4}{=}{Resilience}
& Cost of total unserved energy
& The total cost of energy not delivered to meet load demand due to outages caused by extreme weather. \\
\addlinespace
& SAIDI-MED
& Average total duration of power interruptions per customer per year due to extreme weather events. \\
\addlinespace
& SAIFI-MED
& Average total number of outages experienced per customer per year due to extreme weather events. \\
\addlinespace
& CAIDI-MED
& Average outage duration experienced per customer per year due to extreme weather events. \\
\hline
\addlinespace

Growth
& Growth in distribution plant value
& The change in the total value of distribution assets over the planning period, accounting for depreciation (modeled as a straight line). \\
\hline
\addlinespace

\multirow{3}{=}{Reliability}
& Average capacity headroom of lines
& Average available margin before reaching thermal limits on lines. \\
\addlinespace
& Duration of voltage deviation
& Percentage of time buses exceeded voltage limits ($\pm$5\% of nominal). \\
\addlinespace
& Over-capacity duration
& Number of hours line utilization exceeded 80\% of capacity. \\
\hline
\addlinespace

\multirow{2}{=}{Affordability}
& Cost of total resistive losses
& Total losses due to resistive heating (MWh), aggregated over the grid during the planning period. \\
\addlinespace
& Total investment portfolio cost
& Total cost of the investment plan under consideration. \\
\bottomrule
\end{tabular}
\end{table}

\begin{table}[t!]
\centering
\caption{Parameters for Test Case 1}
\label{test_case_1_values}

\begin{tabular}{L{3.0cm} L{5.0cm}}
\toprule
\textbf{Parameter} & \textbf{Value} \\
\midrule

Total budget & \$20{,}000 \\
\addlinespace
Total time & 1 year \\
\addlinespace
Total technicians & 14 \\
\addlinespace
Simulation start/end & 2016-01-01 -- 2016-12-31 \\
\addlinespace
Possible investments &
9 underground lines upgraded to higher-rated conductors; 4 overhead lines buried. \\
\addlinespace
Outage date for unserved energy calculation & 2016-07-01 \\
\addlinespace
Outage duration for unserved energy calculation & 24 hours \\
\addlinespace
NPV time horizon $T$ & 5 years \\
\addlinespace
Cost per kWh & \$0.20 \\
\addlinespace
Value of lost load & \$5.00/kWh \\
\addlinespace
Rate of return on investments & 8\% \\

\addlinespace
Resources needed to upgrade line & NAYY 4x240SE 0.6/1kV:
\begin{itemize}
  \item 2 days/km
  \item 7 technicians
  \item \$50{,}000/km
\end{itemize} 
\\
\addlinespace

Resources needed to underground line & NAYY 4x240SE 0.6/1kV: 
\begin{itemize}
  \item 8 days/km
  \item 12 technicians
  \item \$175{,}000/km
\end{itemize} \\
\addlinespace

Extreme weather multipliers &
\begin{itemize}
  \item Base rate: 0.0005
  \item Seasonal: winter 2.0, spring 1.0, summer 1.5, fall 1.2
  \item Hourly: night 0.5, morning 1.0, afternoon 1.5, evening 1.2
\end{itemize} \\
\addlinespace

Extreme weather event radius range & $[0.5,\,3.0)$ km \\
\addlinespace

Extreme weather event duration range & $[2.0,\,8.0)$ hours \\

\addlinespace
Probability threshold for line outage &
\begin{itemize}
  \item Overhead line: 0.4
  \item Underground line: 0.05
\end{itemize} \\
\addlinespace

Days required to repair 1 km of line &
\begin{itemize}
  \item Overhead line: 0.5 days
  \item Underground line: 5.0 days
\end{itemize} \\
\addlinespace
NSGA-II parameters & 
\begin{itemize}
    \item Population size: 5
    \item \# of generations: 3
    \item \# of Monte Carlo runs / candidate: 3
\end{itemize} \\

\bottomrule
\end{tabular}
\end{table}

\begin{table}[h]
\centering
\caption{Parameters for Test Case 2}
\label{test_case_2_values}

\begin{tabular}{L{3.0cm} L{5.0cm}}
\toprule
\textbf{Parameter} & \textbf{Value} \\
\midrule

Total budget & \$500{,}000 \\
\addlinespace
Total time & 90 days \\
\addlinespace
Total technicians & 16 \\
\addlinespace
Simulation start/end & 2016-01-01 -- 2016-03-30 \\
\addlinespace
Possible investments &
9 underground lines upgraded to higher-rated conductors; 3 overhead lines buried. \\
\addlinespace
Outage date for unserved energy calculation & 2016-07-01 \\
\addlinespace
Outage duration for unserved energy calculation & 24 hours \\
\addlinespace
NPV time horizon $T$ & 5 years \\
\addlinespace
Cost per kWh & \$0.30 \\
\addlinespace
Value of lost load & \$10.00/kWh \\
\addlinespace
Rate of return on investments & 8\% \\

\addlinespace
Resources needed to upgrade line & NA2XS2Y 1x120 RM/25 12/20 kV:
\begin{itemize}
  \item 7 days/km
  \item 6 technicians
  \item \$135{,}000/km
\end{itemize} 

NA2XS2Y 1x150 RM/25 12/20 kV: 
\begin{itemize}
  \item 8 days/km
  \item 6 technicians
  \item \$150{,}000/km
\end{itemize} 

NA2XS2Y 1x185 RM/25 12/20 kV: 
\begin{itemize}
  \item 9 days/km
  \item 6 technicians
  \item \$180{,}000/km
\end{itemize} 
\\
\addlinespace

Resources needed to underground line & NA2XS2Y 1x95 RM/25 12/20 kV: 
\begin{itemize}
  \item 21 days/km
  \item 8 technicians
  \item \$250{,}000/km
\end{itemize} \\
\addlinespace

Extreme weather multipliers &
\begin{itemize}
  \item Base rate: 0.0005
  \item Seasonal: winter 2.0, spring 1.0, summer 1.5, fall 1.2
  \item Hourly: night 0.5, morning 1.0, afternoon 1.5, evening 1.2
\end{itemize} \\
\addlinespace

Extreme weather event radius range & $[0.5,\,3.0)$ km \\
\addlinespace

Extreme weather event duration range & $[2.0,\,8.0)$ hours \\

\addlinespace
Probability threshold for line outage &
\begin{itemize}
  \item Overhead line: 0.2
  \item Underground line: 0.01
\end{itemize} \\
\addlinespace

Days required to repair 1 km of line &
\begin{itemize}
  \item Overhead line: 0.5 days
  \item Underground line: 5.0 days
\end{itemize} \\

NSGA-II parameters & 
\begin{itemize}
    \item Population size: 8
    \item \# of generations: 5
    \item \# of Monte Carlo runs / candidate: 6
\end{itemize} \\

\bottomrule
\end{tabular}
\end{table}

\begin{figure}[t]
    \centering
    \includegraphics[width=0.5\textwidth]{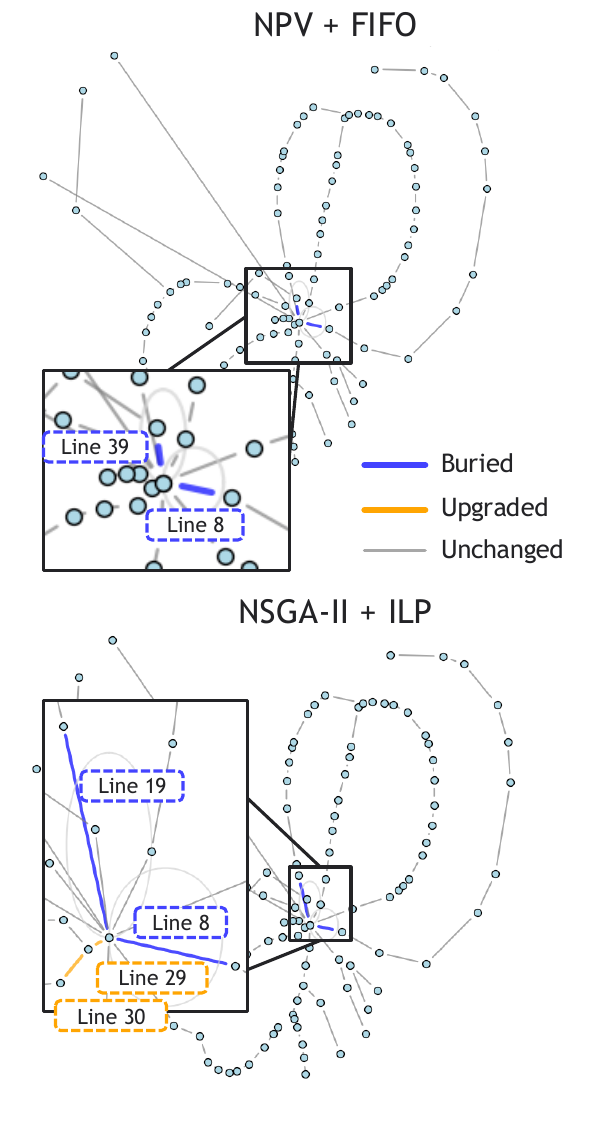}
    \caption{\textbf{Test case 2 results:} \textbf{NPV + FIFO:} Lines 8 and 39 are buried, addressing the two most critical overhead lines on radial branches. \textbf{NSGA-II + ILP:} The optimized portfolio did not include line 39 and instead buried line 19 which was not critical as it was located on a loop branch.}
    \label{fig:test_case_2_results}
\end{figure}

\clearpage
\bibliographystyle{ieeetr}
\bibliography{references}
\balance

\endgroup
\end{document}